\documentclass[11pt,a4paper]{article}
\usepackage[hyperref]{acl2020}
\usepackage{times}
\usepackage{latexsym}
\usepackage{amsmath}

\usepackage{microtype}
\usepackage{algorithm}
\usepackage{algpseudocode}

\usepackage{xcolor}
\usepackage{graphicx}
\aclfinalcopy
\definecolor{cicsls0target}{rgb}{0.7,0.34,0.3}
\definecolor{cicsls0alt1}{rgb}{0.8,0.2,0.27}
\definecolor{cicsls0alt2}{rgb}{0.97,0.03,0.93}
\newcommand\ColorSquare[1]{{\color{#1}\rule{6mm}{6mm}}}

\definecolor{semdial1target}{rgb}{0.71,0.29,0.33}
\definecolor{semdial1alt1}{rgb}{0.73,0.29,0.27}
\definecolor{semdial1alt2}{rgb}{0.67,0.33,0.51}
\definecolor{semdial2target}{rgb}{0.01,0.71,0.99}
\definecolor{semdial2alt1}{rgb}{0.05,0.55,0.95}
\definecolor{semdial2alt2}{rgb}{0.48,0.49,0.52}
\definecolor{semdial3target}{rgb}{0.41,0.59,0.52}
\definecolor{semdial3alt1}{rgb}{0.81,0.55,0.19}
\definecolor{semdial3alt2}{rgb}{0.22,0.78,0.45}
\definecolor{semdial4target}{rgb}{0.59,0.2,0.8}
\definecolor{semdial4alt1}{rgb}{0.68,0.32,0.65}
\definecolor{semdial4alt2}{rgb}{0.43,0.58,0.42}
\definecolor{semdial5target}{rgb}{0.71,0.4,0.29}
\definecolor{semdial5alt1}{rgb}{0.99,0.39,0.01}
\definecolor{semdial5alt2}{rgb}{0.84,0.5,0.16}

\definecolor{semdial-baseline1-p1}{rgb}{0.98, 0.02, 0.91}
\definecolor{semdial-baseline1-p2}{rgb}{0.74, 0.26, 0.71}
\definecolor{semdial-baseline1-p3}{rgb}{0.69, 0.07, 0.93}

\newcommand\Tstrut{\rule{0pt}{2.6ex}}         
\newcommand\Bstrut{\rule[-0.9ex]{0pt}{0pt}}   
\aclfinalcopy

\title{Discourse Coherence, Reference Grounding and Goal Oriented Dialogue}

\author{
Baber Khalid$^\dagger$ 
Malihe Alikhani$^{\dagger*}$ 
Michael Fellner$^*$ 
Brian McMahan$^\dagger$ 
Matthew Stone$^{\dagger*}$\\
  $^\dagger$Rutgers Computer Science Department \\
  $^*$Rutgers Center for Cognitive Science \\
\texttt{\{firstname.lastname\}@rutgers.edu}}

\date{}

\begin{document}
\maketitle
\begin{abstract}
Prior approaches to realizing mixed-initiative human--computer referential communication have adopted information-state or collaborative problem-solving approaches.  In this paper, we argue for a new approach, inspired by coherence-based models of discourse such as SDRT \cite{asher-lascarides:2003a}, in which utterances attach to an evolving discourse structure and the associated knowledge graph of speaker commitments serves as an interface to real-world reasoning and conversational strategy.  As first steps towards implementing the approach, we describe a simple dialogue system in a referential communication domain that accumulates constraints across discourse, interprets them using a learned probabilistic model, and plans clarification using reinforcement learning.
\end{abstract}

\section{Introduction}

Referential communication tasks have long served benchmark problems for human--computer dialogue systems \cite{heeman-hirst-1995-collaborating}, and they remain a topic of current investigation \cite{monroe-etal-2017-colors}.  One reason is the flexibility and robustness of human joint activity in referential communication \cite{clark_hh-wilkesgibbs:1986a}.  Another is the complexity of bootstrapping and reconciling wide-coverage models of semantics in dialogue, on the one hand, and effective models of collaboration under uncertainty, on the other. 

Computational work on dialogue has generally integrated collaboration with semantic processing using act-based models \cite{traum1994computational,larsson/traum:nle00,rich:ai01,devault:2008,galescu-etal-2018-cogent}.  On these approaches, problem-solving actions are recognized indirectly from utterance content using intention recognition, and thereby integrated into a hierarchical model of the dialogue's open goals and plans.   While these approaches are powerful and successful, it remains challenging to scale up the necessary models of dialogue structure and intention recognition.  For example, system builders have to map out an inventory of transitions that interlocutors can use to update the problem-solving state and have to describe how those transitions are signaled by utterances with specific kinds of form and meaning.

This paper advocates an alternative approach to link the theory of collaborative activity to the practice of dialogue system design, inspired by coherence-based approaches to dialogue semantics such as SDRT \cite{asher-lascarides:2003a,lascarides:asher:2009}.  On a coherence-based approach, collaboration plays out at the level of the discourse as a whole, rather than at the level of the individual utterance.   In particular, semantic interpretation attaches utterance content directly into an evolving discourse structure, from which the commitments of interlocutors can be derived as a knowledge graph; it is coherence-based interpretation, rather than plan recognition, that builds a hierarchical structure for the dialogue.  Collaborative reasoning comes in later, to connect the coherent dialogue content to its implications for interlocutors' goals, plans and problem solving.  In particular, semantic evaluation operations, trained by reinforcement learning or supervised learning from generic resources, ground the knowledge graph in the task context and underwrite system strategy.  

We begin this paper with a high-level perspective on this approach, highlighting the relevance of discourse-level interpretation for collaborative reasoning and sketching the possible benefits of eliminating step-by-step models of the task contributions of utterances in favor of holistic task-based evaluations of discourse content.  As a first step in implementing this approach, we describe a proof-of-concept system that distinguishes arbitrary color patches from alternatives, as in \citet{monroe-etal-2017-colors}.  We emphasize our extensive (but not complete) use of self-supervised and reinforcement learning methods in constructing the system, and report preliminary results showing the effectiveness of a baseline version of the system in interaction with crowd workers.  This experience suggests that a coherence-based approach offers a promising alternative direction for systematic efforts to scale up computational approaches to flexible, mixed-initiative dialogue.

\section{Motivation}
\label{motivation-sec}

The key linguistic insight of our work is that referential communication depends on semantic grounding operations that apply at the level of the discourse as a whole, operations that draw not only on the explicit commitments of interlocutors but also on those signalled implicitly.
We begin by showing how these insights are captured by coherence approaches to discourse representation, particularly SDRT \cite{asher-lascarides:2003a,lascarides:asher:2009}, and motivating their importance for referential communication tasks.
We argue further that coherence-based approaches create novel, and potentially attractive engineering trade-offs in capturing these phenomena in interactive systems.

\begin{figure}[t!]
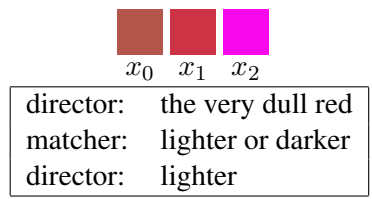

\centering
\begin{tabular}{l}
    \multicolumn{1}{c}{\rule{0pt}{6ex}
        \shortstack{\ColorSquare{cicsls0target} \\ $x_0$}
        \shortstack{\ColorSquare{cicsls0alt1} \\ $x_1$}
        \shortstack{\ColorSquare{cicsls0alt2} \\ $x_2$}
    } \\
    \begin{tabular}{|ll|}
        \hline
        director: & the very dull red \\
        matcher: & lighter or darker \\
        director: & lighter\\
        \hline
    \end{tabular} 
 \end{tabular}
\caption{An example from the Colors in Context (CIC) dataset \cite{monroe-etal-2017-colors} of the director and matcher coordinating so that the matcher can click on the correct color patch ($x_0$).}
\label{fig:cic-example1}
\end{figure}

As the empirical starting point for our work, we use the Colors in Context (CIC) dataset of \citet{monroe-etal-2017-colors}.  
They asked participants to talk about items in a visual display using a free-form chat interface.  
On each round of interaction, one human subject, designated the director, was provided privately with a target item from a randomized display and tasked with instructing the other human subject, designated the matcher, to click on the correct item.  
Examples from this dataset are shown in Figure~\ref{fig:cic-example1}--\ref{fig:cic-example4}. 
In line with the longstanding findings of empirical work on referential communication since \citet{clark_hh-wilkesgibbs:1986a}, both director and matcher assume responsibility in these dialogues to make sure that the team members arrive at a mutually satisfactory description of the target patch and complete the task successfully.

\subsection{Coherence}

Coherence theory \cite{hobbs_jr:1979a,hobbs1985coherence,kehler:coherence01,asher-lascarides:2003a} characterizes conversation through a small number of qualitative relations that link each utterance to an evolving representation of the dialogue.  Linking an utterance to a related segment can create a sibling (coordination) or a child (subordination), generating a hierarchical structure.  In some approaches to discourse coherence, including SDRT, contributions can have multiple attachments, so discourse structure is a complex graph, not just a tree.%
\footnote{\citet{hobbs1985coherence} and \citet{kehler:coherence01} seem to focus on tree-like discourse structures, for example.}
This seems to be particularly important for situated conversation \cite{hunter:etal:2018}, including talk about visual contexts like the displays of the CIC dataset.

Coherence relations come in a number of qualitatively different varieties.  Some relations signal inferences at the level of domain entities: \emph{Narrative} is a coordinating relation involving descriptions of successive events; \emph{Expansion} is a subordinating relation involving the extended description of a single entity.  Some relations are best described in terms of propositions: \emph{Parallel} is a coordinating relation that presents two propositions as analogous; \emph{Explanation} is a subordinating relation connecting a fact to the generalization that explains it.  A final class of relations involve meta-level relationships that involve the goals or circumstances of utterances.  \emph{Correction} in dialogue is a coordinating relationship of this kind; \emph{Clarification Request} is a subordinating one.   Coherence theories postulate substantial divergences between these three different classes of relations.  However, relations of any kind suffice to make the conversation coherent and to underwrite inferences about the content of the conversation as a whole, including the commitments of the different speakers. 

\citet{schlangen-lascarides-2003-interpretation} argue for the suitability of using coherence to characterize the interpretation of the fragmentary utterances---such as those in Figure~\ref{fig:cic-example1}---that are ubiquitous in referential communication, and \citet{stone:lascarides:2010} show how representations of coherence and speaker commitment can underwrite probabilistic inferences about interlocutors' shared understanding in dialogue.   We begin by reviewing, extending and deepening those formal accounts.  Alternative approaches to these issues are of course possible \cite[among others]{ginzburg:2012}, and we are certainly not in a position to argue against other approaches at this stage of our work.  However, we do argue that coherence theory offers an elegant, integrated framework that addresses the structural, semantic and inferential challenges of referential communication in a way that extends to other kinds of dialogue interaction as well.

\subsection{Coherence and Dialogue Structure}

Coherence theory offers a simple and natural way to represent the \textsc{structure} of interactions such as that in Figure~\ref{fig:cic-example1}.  Let's look at this interaction in detail.  It's useful to segment the utterances into basic moves that establish reference, commit to propositions and contribute open questions.  This lets us describe the dialogue as an action $u_1$ that refers to the target with the description \emph{the very dull red}; a question $u_2$ whether the target is \emph{lighter} and a question $u_3$ whether the target is \emph{darker}, combined via an alternation relation into segment $\sigma_4$ contributing a complex question; and a contribution $u_5$ that the target is \emph{lighter}.  These fit together into a hierarchical structure:
\begin{center}
\begin{tabular}{cl}
    $u_1$ & discourse initial \\
    $\sigma_4$ & $\textit{Alternative}(u_2, u_3)$ \\
           & subordinate to $u_1$ \\
           & $\textit{Clarification}(u_1, \sigma_4)$ \\
    $u_5$ & coordinate to $\sigma_4$ \\
          & $\textit{Answer}(\sigma_4, u_5)$ \\
          & and $\textit{Answer}(u_2, u_5)$
\end{tabular}    
\end{center}
The hierarchical structure here follows from the definitions of the relations involved---\emph{Clarification}, \emph{Alternative}, \emph{Answer}---and is built incrementally as moves attach to the discourse. Nevertheless, it supports the key inferences you'd expect for a hierarchical discourse structure, including the possibility for moves to attach high and return to an ongoing topic after a subsidiary inquiry, and the incoherence that interlocutors typically feel when topics are disjointed or interleaved.  Moreover, it delivers these inferences as part of general model of dialogue that can be relevant for other tasks such as planning successful dialogue.  For example, recognizing terms of the matcher's question \emph{lighter or darker} are parallel alternatives (which might be specified or learned as a typical feature of effective clarification questions). 

\subsection{Coherence and Semantic Reference}

Coherence theory also offers a sophisticated approach to the \textsc{semantics} of referential communication dialogues.
Like all referential communication, the dialogue of Figure~\ref{fig:cic-example1} is about its contextual situation: the director's utterances work to identify patch $x_0$.  It's common to think of such situated reference in terms of the compositional semantics and pragmatics of individual expressions, as in \citeauthor{kripke:1977a2}'s \citeyearpar{kripke:1977a2} philosophical account of speaker reference or the words-as-classifiers paradigm of \citet{kennington-schlangen-2015-simple}.  However, researchers as early as \citet{luperfoy-1992-representation} have argued that the management of situated reference in the presence of anaphoric relations and partial information requires representing and reasoning about the content of the discourse as a whole.  

\begin{figure}[b!]
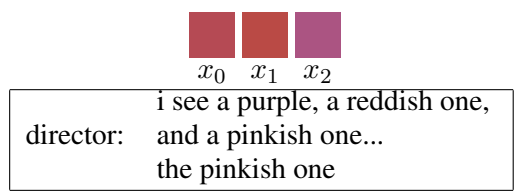

\centering
\begin{tabular}{c}
    \multicolumn{1}{c}{\rule{0pt}{6ex}
        \shortstack{\ColorSquare{semdial1target} \\ $x_0$}
        \shortstack{\ColorSquare{semdial1alt1} \\ $x_1$}
        \shortstack{\ColorSquare{semdial1alt2} \\ $x_2$}
    } \\
    \begin{tabular}{|ll|} 
    \hline
        \shortstack[l]{\Tstrut director:\\\rule{0pt}{2ex} } & 
        \shortstack[l]{
            i see a purple, a reddish one, \\
            and a pinkish one... \\
            the pinkish one
        }  \Bstrut \\
     \hline
     \end{tabular}
\end{tabular}
\caption{Semantic grounding needs to apply at the level of the discourse to handle anaphoric relations.}
\label{fig:cic-example2}
\end{figure}

Figure~\ref{fig:cic-example2} shows how the problem comes up in the CIC dataset.  In cases such as that shown in Figure~\ref{fig:cic-example2}, directors frame difficult reference problems not just by describing the target but also by giving contrasting descriptions of relevant alternatives.  Subsequent utterance can identify objects in context by exploiting these established expressions. 

Coherence theory might describe this dialogue in a fine-grained way via the composition of four moves by the director (for now we ignore the contribution of \emph{i see}).  The director begins with three descriptions that introduce discourse referents into an evolving discourse model: Such inferences are the bread-and-butter of the dynamic approaches to discourse semantics on which coherence approaches such as SDRT are based; see \citet{dekker:2012}.  We will represent this as three fragments $u_1$, $u_2$ and $u_3$, related by a \emph{Parallel} relation into a segment $\sigma_4$.  This segment is subordinate to the follow up \emph{the pinkish one} and attaches to it via a \emph{Background} relation; \emph{the pinkish one} also attaches to $u_3$ by a \emph{Presupposition Resolution} relation through which the definite noun phrase is resolved anaphorically to the discourse referent introduced by $u_3$.  The result is an integrated representation of the speaker's commitments in the conversation that features just three discourse referents, each of which is constrained to pick out a different entity in the visual scene, and one of which is designated as giving the target referent for the task.

Note how coreference constraints, disjoint reference constraints, and visual grounding work together at the level of the discourse in interpreting the director's contribution.  Evidently, the description \emph{pinkish one} on its own leaves substantial uncertainty about the intended referent; indeed, $x_2$ might be a more likely match than $x_0$.  However, the implicit constraint that the target can be distinguished from another patch that's \emph{purple} and another that's \emph{reddish} significantly reduces the possibility of misinterpretation.  Any constraint-satisfaction approach to ambiguity resolution will handle this, whether logical \cite{mellish:descriptions} or probabilistic \cite{10.1145/1088463.1088489}.  But what's to prevent the original uncertainties to resurface when we interpret a new token of the description \emph{the pinkish one}? The natural way is to follow the coherence-based account, and invoke a discourse-level inference that interprets a second occurrence as dependent on (and therefore referentially identical to) the first, in virtue of an anaphoric link.  

\subsection{Coherence and Common Ground}

Coherence itself is vital for calculating an integrated representation of dialogue content \cite{lascarides:asher:2009}.  Content is often contributed and acknowledged implicitly, in virtue of the coherence relations that attach utterances to the discourse.  This is easily regimented in coherence theory.  \citet{lascarides:asher:2009} give a formal model that exploits coherence to identify the commitments that interloctors have made over the course of a coherent discourse.  Such calculations can be crucial to tracking mutual understanding in dialogue \cite{stone:lascarides:2010}, as we can see even in the CIC domain.

\begin{figure}[b!]
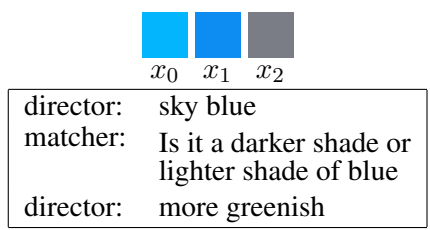

\centering
\begin{tabular}{c}
    \multicolumn{1}{c}{
        \shortstack{\ColorSquare{semdial2target} \\ $x_0$}
        \shortstack{\ColorSquare{semdial2alt1} \\ $x_1$}
        \shortstack{\ColorSquare{semdial2alt2} \\ $x_2$}
    } \\
    \begin{tabular}{|ll|} 
    \hline
        director: & sky blue \\
        \shortstack[l]{ matcher:\\\rule{0pt}{2ex}} 
        & 
            \shortstack[l]{
                Is it a darker shade or  \\ 
                lighter shade of blue
            }\\
        director: & more greenish \\
     \hline
     \end{tabular}
\end{tabular}     
\caption{Clarification questions implicate the matcher's implicit acceptance of the possibility of grounding the director's description in the visual scene.}
\label{fig:cic-example3}
\end{figure}

Take clarification requests, as in Figures~\ref{fig:cic-example1} and~\ref{fig:cic-example3}.  Even though clarification requests flag particular aspects of interpretation as incomplete, they implicitly accept the descriptions they target as appropriate to the visual scene.  Thus, the matcher of Figure~\ref{fig:cic-example1} seems willing to accept $x_0$ or $x_1$ as a very dull red; that of Figure~\ref{fig:cic-example3} seems willing to accept $x_0$ or $x_1$ as a sky blue.  In particular, the dialogue of Figure~\ref{fig:cic-example3} illustrates how these implicit commitments can inform interlocutors' subsequent coordination: the utterance \emph{more greenish} must be understood to select from the sky blue candidates that have been mutually agreed; $x_3$ may well be the \emph{more greenish} patch in the display as a whole.

\begin{figure}[b!]
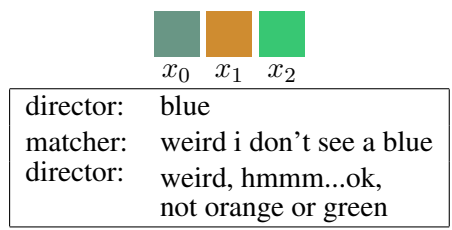

\centering
\begin{tabular}{l}
    \multicolumn{1}{c}{
        \shortstack{\ColorSquare{semdial3target} \\ $x_0$}
        \shortstack{\ColorSquare{semdial3alt1} \\ $x_1$}
        \shortstack{\ColorSquare{semdial3alt2} \\ $x_2$}
    } \\
    \begin{tabular}{|ll|} 
    \hline
        director: & blue \\
        matcher: & weird i don't see a blue \\
            \shortstack[l]{ director:\\\rule{0pt}{2ex}} 
            & 
            \shortstack[l]{
                weird, hmmm...ok, \\
                not orange or green 
            }
        \\
     \hline
     \end{tabular}
\end{tabular}     
\caption{Implicit commitments contrast with outright rejections.  Here, the director's followup strategy of excluding distractors seems uninformed by expectations or conclusions the matcher might have drawn from the initial utterance.}
\label{fig:cic-example4}
\end{figure}

Figure~\ref{fig:cic-example4} shows what happens when no such implicit commitments can be inferred.  Since the matcher's response completely rejects the director's initial description, the director proceeds with an alternative referential strategy---exclusion of the two other patches---that stands independently of the content contributed by the initial utterance.

\subsection{Implementing Coherence}

In this paper, we explore coherence not just as a theoretical model for dialogue but as an architecture for implementing dialogue systems.   Our overarching vision is that utterances in dialogue should be interpreted with a grammatically-specified meaning that's resolved against the linguistic context and attached to an evolving discourse structure.  This meaning need not specify abstract updates to the state of the interaction, such as grounding acts or problem-solving acts; it just specifies what's required to track the content of the discourse.  It is discourse-level inferences that link speakers' commitments to the real-world task context and underwrite the prediction of the outcomes of conversational strategies based on system experience.

We can contrast this architecture with problem-solving approaches that capture semantic grounding and contributions to collaborative activity by reasoning about individual utterances.  Compared to a discourse-level approach, an utterance-level architecture is likely to require a finer-grained integration of system capabilities, correspondingly finer-grained data resources---and, given current methods---substantially more handcrafted representation and data annotation.

\section{System Architecture}
\label{system}

Realizing the vision we have sketched in Section~\ref{motivation-sec} represents a long-term project for dialogue systems design.  In particular, it requires us to adapt linguistic processing models so that utterance interpretations contribute to a structured logical form for conversation; it requires us to adapt reasoning and planning frameworks so they apply at the level of discourse content.  Will it be possible to make the changes involved?

We think it will.  In the rest of this paper, we argue for the practical feasibility of a coherence-based approach to dialogue architecture, using our initial experience implementing a dialogue system using the design principles articulated in Section~\ref{motivation-sec} that successfully plays the referential game investigated by \citet{monroe-etal-2017-colors}.
As described in Section~\ref{sec:grammar}, we used the development dialogues of the CIC dataset to design linguistic processing models that interpret utterances as coherent updates to discourse-level representations of logical form.  Then, as described in Section~\ref{composite_model}, we combined machine-learned models of lexical semantics and referential interpretation with rule-based models of compositional semantics to characterize the referential ambiguities of these discourse representations.  Finally, as described in Section~\ref{sec:rl}, we characterize system abilities in terms of a set of high-level strategies for contributing to conversation, and implement a reinforcement-learning approach to selecting those strategies based on their discourse-level effects.  These three interacting reasoning models work together to operationalize our approach to referential communication as an empirically-based, discourse-level collaborative process.


\subsection{Handling Utterances}
\label{sec:grammar}

Our approach to utterance interpretation is based on a handcrafted context free grammar that compositionally interprets utterances in terms of corresponding formal contributions to dialogue. 

The basic units of content in the system are color predications drawn from the XKCD lexicon presented in \citet{mcmahan-stone-2015-bayesian}.  Using the inference methods described in Section~\ref{composite_model}, we can interpret these predications descriptively, simply as saying what a color is like, or referentially, as identifying the unique patch that fits its content.  Compositional operations allow these predications to be combined by logical operations into complex descriptions.  Contributions to logical form can introduce discourse referents and characterize them in terms of descriptions; these contributions are defined recursively in terms of coherence relations, which are attached into the overarching discourse structure.  Given a discourse structure, we can traverse it to determine the commitments of the director, which involve a set of discourse referents, a sequence of descriptive characterizations of those discourse referents, and inequalities among discourse referents.  One of these discourse referents is distinguished as representing the target of the referential communication task.

Our logical form language offers the following moves at the level of discourse: \textit{Identify}, \textit{Distinguish}, \textit{Ask\_Clarification}, \textit{Confirmation}, \textit{Rejection}.
$\textit{Identify}(x,D)$ commits the speaker that description $D$ identifies discourse referent $x$; for example, \emph{the green one} corresponds to a move \textit{Identify(T,green)} (for the distinguished target $T$). $\textit{Distinguish}(x,y)$ commits the speaker that discourse referents $x$ and $y$ are different; for example \emph{not the dark blue} corresponds to   two related moves $\textit{Distinguish(T, A)}; \textit{Identify(A,dark blue)}$.  In this simple implementation, \textit{Ask\_Clarification} proposes a description of the target object, to which nobody is committed. \textit{Confirmation} (\emph{Rejection}) commit the speaker that this description fits (does not fit) the target.

Every utterance contributes a node which attaches to existing node(s) in the discourse graph. Initially, the discourse graph consists of a single node introducing the target discourse referent.  With each new description of target by the director we add a new node in the graph which attaches to most recent node by the director. Whenever a new clarification question comes, it adds a new node in the graph which attaches to the initial node.  A confirmation or rejection attaches to the most recent node representing a clarification question and the most recent node by the director.

We handle the idiosyncratic syntax and telegraphic structure of color words, acknowledgments and other frequent contributions in CIC utterances using a domain-specific syntactic specification, developed by analyzing the development set of the CIC corpus.\footnote{For example, when analyzing the data, it became apparent that the fact that whether a given occurrence of color name figured syntactically as either a noun or an adjective had no bearing on the meaning derived for the utterance as a whole.}  We illustrate our approach with a small excerpt from our context-free grammar (CFG) in Table~\ref{tab:cfg}. 

\begin{table}[H]
    \centering
    \begin{tabular}{|c|c|c|}
    \hline
     Category & \multicolumn{2}{|c|}{Expansions} \\
     \hline\hline
     $\langle S\rangle$ & $CP$ & $NegP$\\
     \hline
     $\langle CP\rangle$ & $ADJ : CLR$ & $CLR$\\
     \hline
     $\langle ADJ\rangle$ & $grassy$ & $super$\\
     \hline
     $\langle NegP\rangle$ & $NEG : CP$ & $NEG : ADJ$\\
     \hline
     $\langle NEG\rangle$ & not &\\
     \hline
     $\langle CLR\rangle$ & \multicolumn{2}{|c|}{Lux Lexicon}\\
     \hline
    \end{tabular}    
    \caption{A small portion of the CFG developed for our dialogue system's interpretation engine (`:' denotes string concatenation).}
    \label{tab:cfg}
\end{table}


We extend our basic grammar to a probabilistic-CFG (PCFG) by automatically inducing the relevant weights from the CIC train split.  In particular, we obtain all parse trees of the 15285 utterances in the CIC train split, determine the commitments the parse trees contribute to the dialogue, and then weight the trees by the probability that those commitments converge on the actual target of the interaction using the models of Section~\ref{composite_model}.  The resulting weighted expansions are used to estimate the probabilities associated with our CFG rules.

We use Earley's algorithm for CFG parsing and A* search to obtain the most likely parse tree given a PCFG.  To handle incomplete parses, we used the \textit{first-set} of grammar variables to find the next position where parser could resume to find a sub-parse tree.

Table~\ref{tab:parse_coverage} shows the coverage of our parser over the train split of CIC dataset. The performance shows that our lexicon and grammar rules were able to capture the vast majority of words and structures used by human interlocutors in this setting.

\begin{table}[ht!]
    \centering
    \begin{tabular}{|p{3.8cm}|r|}
    \hline
     Parse Status & Rate \\
     \hline
     Complete Parse    &       $.8000$\\
     \hline
     One NOPP & $.1329$\\
     \hline
     Two NOPPs & $.0353$\\
     \hline
     Three or More NOPPs & $.0020$\\
     \hline
     No Parse & $.0283$\\
     \hline
    \end{tabular}
    \caption{Parsing coverage on CIC training data, including complete parses, non-overlapping partial parses spanning segments of the input (NOPPs), and complete parse failure.}
    \label{tab:parse_coverage}
\end{table}

To evaluate parser accuracy, we looked at the first utterances in each dialogue of the CIC test set, and focused on just those cases where the parser returned at least one complete parse.  We took the most likely parse tree and computed its semantic interpretation and evaluation.  If the actual target of that interaction had the largest posterior probability of the candidate patches, we counted this as a successful item.  Overall, the success rate was 88.9\%. Even though this evaluation method does not confirm the correctness of the parse itself, we believe that it provides strong evidence for the broad accuracy of our parser.

\subsection{Semantic Evaluation}
\label{composite_model}

Given these representations of discourse content, we define a semantic evaluation operation to link that discourse content to the visual world.  The input is the structure of speaker commitments.  The output is a posterior distribution over candidate color patches for the target. (Although in principle our logical form language allows us to compute posteriors over color patches for any of the discourse referents evoked in the dialogue, we focus here on the typical case where we only need to track the distinguished discourse referent for the target.)  In other words, sematnic evaluation defines a probability distribution 
\[
P(x_i | w^t, w^{t-1},...,w^0, C)
\]
where $w^t$ represents the descriptive constraint on the target associated with the $t$th speaker commitment, $x_i$ ranges over the candidate color patches in the scene, and $C$ indicates the dependency of the calculation on a specific visual context, in this case given by three color patches.

For atomic descriptions from the XKCD lexicon, we use the models developed in \newcite{Brian-Matthew-SigDial}, which are available at \url{https://go.rutgers.edu/ugycm1b0}.  These models are learned in two steps directly from naturally-occurring data.  The first step is to train a model of color term meaning from Randall Munroe's crowdsourced collection of free text descriptions of color patches, as curated by \newcite{mcmahan-stone-2015-bayesian}.  The next step is to use cognitive modeling and a latent variable model to estimate the prevalence of different speaker strategies for using color meanings to construct identifying descriptions from the CIC dataset. The result is a probability distribution 
\[ 
P(w_k | x_i, C) 
\]
describing the likelihood that a human speaker will use the term $w_k$ to identify a target $x_i$ in the context $C$.  This is readily converted to $P(x_i | w_k, C)$ by Bayes's theorem (on the assumption that each of the targets is equally likely \emph{a priori}).

To extend atomic probabilities to complex predications, we use logical definitions.  For example, for a conjoined predication $w^{t1} \wedge w^{t2}$ we have:
\[
P(x^i | w^{t1} \wedge w^{t2}) = P(x^i | w^{t1}) P(x_i | w^{t2})
\]
For disjunction, we get $P(x_i | w^{t1} \vee w^{t2}) =$
\[
1 - ( (1 - P(x_i | w^{t1})) (1 - P(x_i | w^{t2}))) 
\]
by de Morgan's rule.

Finally, we treat the commitments of the speaker as providing independent evidence about the identity of the target.  Again under the assumption that all targets are equally likely \emph{a priori}, this gives
\[
P(x_i | w^t, w^{t-1}, ... , w^0, C) 
    \propto \prod_{i=0}^{t}P(x_i | w^i, C) 
\]

\subsection{Learning Data-Driven Decisions}
\label{sec:rl}

Act-based approaches to collaboration plan contributions to conversation based on formal models of the preconditions and effects of moves in conversation: the system can reason backward from conversational goals to the actions it might use to achieve them.  In our approach, the formal effects of conversational actions are defined exclusively by the commitments of interlocutors, as represented in discourse structure, which offers a comparatively thin notion of the dynamics and goals of conversation.  We therefore need a different approach to reason about what actions to take next.  Our approach is reinforcement learning, which is based on characterizing the outcomes that count as successes in dialogue and anticipating the course of conversational interaction.

A reinforcement learner represents its decision making as a Markov Decision Process (MDP), defined by states $S$, actions $A$, a transition function $T$, rewards $R$, and discount factor $\gamma$.  It iteratively improves its estimates of the expected payoffs for an action $a$ in a state $s$ based on experience, and ultimately selects the action whose estimated payoff is the highest.  In particular, using data from the Colors in Context dataset, we train a reinforcement learner to approximate the state-action value function under a decision-making policy $\pi$, otherwise known as the Q-function, which is given as:
\begin{center}
$Q^{\pi}(s, a) = r(a) + \gamma * E[ max_{a'\in A}Q^{\pi}(s', a') ]$ 
\end{center}
We use a feature-based representation of the dialogue state and function approximation to drive a uniform, general estimate of the $Q$-function.

Developing this approach involves establishing formal models for the actions, reward, state and effects of conversational actions.
Our system chooses from two possible actions in each turn, \textit{Ask\_Clarification} and \textit{Select}.  The description associated with the \textit{Ask\_Clarification} move is determined by the referential goal communication model of \newcite{Brian-Matthew-SigDial} applied to the most likely candidate referent; we generate the most likely expression not previously used in the dialogue. The \textit{Select} action, meanwhile, picks the most likely color patch $\mu$ under the posterior defined in Section~\ref{composite_model}.

We represent the state $s^t$ using features so as to record the history of the dialogue and the current information available about the target:
\begin{gather}
s_i = \{\forall x_i \; P(x_i | w^t, ... , w^0, C) : [a^{0}, ... , a^{t}]\} \nonumber
\end{gather}
The prior actions in the dialogue are represented as a vector using a one-hot encoding.

The reward function $r$ is designed to create a tradeoff between dialogue length and task success; good dialogues are both short and effective.
In particular, if the correct target is $t$
\[
\begin{array}{l}
r(\textit{Select})=0.3 \; \textrm{if} \; \mu=t \nonumber \\
r(\textit{Select})=-0.5 \; \textrm{if} \;\mu\neq t \nonumber \\
r(\textit{Ask\_Clarification})=-0.1 \nonumber 
\end{array}
\]
After 15 turns the dialogue ends in failure with a reward of $-1$ and no further actions are possible.
 
In order to anticipate the long-term effects of system choices, we model the responses that the system elicits.  This is known as a user simulation.  Initially, the simulation identifies the target using a sampled description controlled by a parameter $P_x$ which represents the probability the user will succeed in generating an identifying description.  With probability $P_x$, the user presents a description that's merely true of the target patch using the semantic model of \newcite{Brian-Matthew-SigDial}, but with probability $1-P_x$ the user's description is sampled from their model of identifying descriptions.  If the system clarifies, the user simulation confirms if the most likely referent of the distribution is the target, it responds with a \textit{Confirmation}; otherwise it responds with a \textit{Rejection} and a new sampled description.

\section{Analysis and Results}
\label{eval}

In Section~\ref{system}, we presented our initial approach to characterizing semantic evaluation and collaborative reasoning at the level of discourse structure.  We've seen that we can characterize utterances effectively in this framework.  Now we show further that the semantic evaluation operations and collaborative reasoning models are effective.

\subsection{Human Evaluation}
\label{sec:human_eval}

We ran a human evaluation experiment with a simple rule-based baseline system that uses the semantic evaluation models of Section~\ref{composite_model} to decide on dialogue strategy.  This baseline system went for a \textit{Select} action whenever the probability of most likely referent was greater than 0.95 and chose to ask a clarification request otherwise.

\paragraph{Protocol.}

This study was conducted with the approval of
our human subjects review committee. We recruited 50 subjects through Amazon Mechanical Turk. Participants were all US citizens, gave
written consent, and were compensated at an estimated rate of USD 15 an hour. Each subject played 4 games with the system out of which two were selected from the \textsc{close} condition of \newcite{monroe-etal-2017-colors}, one from their \textsc{far} condition and one from their \textsc{split} condition. (According to our sensitivity power analysis, with a sample size of 176 conversations we would be able to detect effect sizes as small
as 0.820 with a power and significance level of
90\%.) After each trial, subjects were asked to rate the performance of the system on a scale of 0 to 5 and leave us feedback.   

\paragraph{Results.}

The overall rate of successful trials is 72.6\% and the average overall rating of the system is 3.7.  In understanding the effectiveness of discourse-level semantic operations, we are particularly interested in the course of dialogues where information accumulates over time.  There are actually 92 cases where the system simply chooses to select on the first turn, picking the correct target in 76 of them (over 82\%).  In the 123 trials where the system asks for clarification after the first turn, just 78 led to successful identification---a rather disappointing success rate of just over 63\%.  However, the problem seems to be the users' interpretation of system utterances, not the system's integration of content across the discourse.  System clarification questions include descriptions that the utterance-level model of \newcite{Brian-Matthew-SigDial} predicts will identify a particular color patch.  If that patch is in fact the target of this round, we'd expect users to confirm; if that patch is not the target of this round, we'd expect users to reject.  However, users' answers align with the answers we would expect them to give based on the true target only 48\% of the time.  Truly effective systems will need to ask clarification questions that lead to more reliable answers.

\begin{figure}[t!]
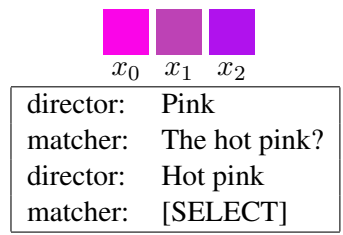

\centering
\begin{tabular}{l}
    \multicolumn{1}{c}{\rule{0pt}{6ex}
        \shortstack{\ColorSquare{semdial-baseline1-p1} \\ $x_0$}
        \shortstack{\ColorSquare{semdial-baseline1-p2} \\ $x_1$}
        \shortstack{\ColorSquare{semdial-baseline1-p3} \\ $x_2$}
    } \\
    \begin{tabular}{|ll|}
        \hline
        director: & Pink \\
        matcher: & The hot pink? \\
        director: & Hot pink \\
        matcher: & [SELECT] \\
        \hline
    \end{tabular} 
 \end{tabular}
\caption{An example conversation of a human director interacting with our system using a baseline matcher policy.}
\label{RL:task_example}
\end{figure}

Nevertheless, Figure~\ref{RL:task_example} shows an example conversation of the user with our system that shows how the use of clarification does sometimes lead to a natural and effective conversation in this task. We are releasing the data collected for this evaluation.\footnote{https://github.com/baber-sos/Discourse-Coherence-Reference-Grounding-and-Goal-Oriented-Dialogue}

\subsection{RL Experiments}

We ran computer experiments in our RL framework to demonstrate that our approach converges to sensible policies for contributing to a collaborative interaction.

\paragraph{Method.}
We estimated action outcomes using the deep $Q$ learning (DQN) algorithm of \citet{mnih2013atari}. The $Q$ function is approximated using linear function $\theta_p$ called \textsc{policy net}. Another linear function $\theta_t$ also known as \textsc{target net} is used to predict the future value of a state $s$. A delta $\delta$ is calculated as below and the gradient loss of this error is back-propagated to learn the $Q$ function.
\begin{gather}
\delta = Q^{\pi}_{\theta_p}(s, a) - (r(a) + \gamma * max_{a' \in A}Q^{\pi}_{\theta_t}(s', a')) \nonumber
\end{gather}

Training is done by sampling a mini-batch of size $24$ from an experience replay memory of size $200$ with a circular buffer, calculating the $\delta$ using the {Huber} loss and then back-propagating using the {Adam} optimizer. Also, instead of updating $\theta_t$ at each iteration using back-propagation, it gets updated by copying the weights of $\theta_p$ every \textit{20} iterations. We set the discount factor $\gamma=1$, learning rate $\alpha = 0.001$ and weight decay parameter $\lambda=0.01$ during the training process.We train our RL framework for 4000 dialogue simulations on color samples from CIC train split. The model is then tested on 400 simulations over unseen color samples.


\paragraph{Results.}

\begin{figure}[h!]
  \includegraphics[width=\columnwidth]{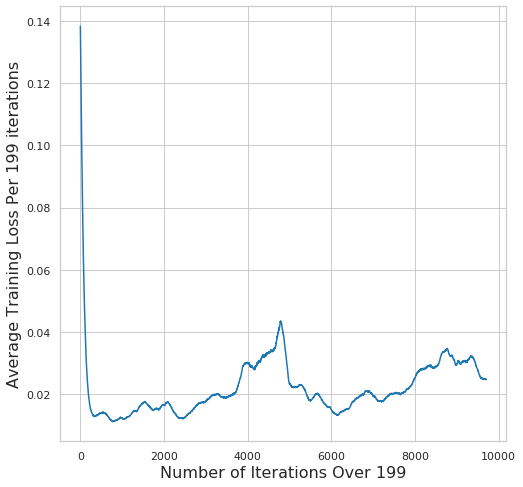}
  \caption{Progression of training loss over 4000 simulated conversations with color samples for CIC dataset train split. }
  \label{fig:trainLoss}
\end{figure}

\begin{figure}
  \includegraphics[width=\columnwidth]{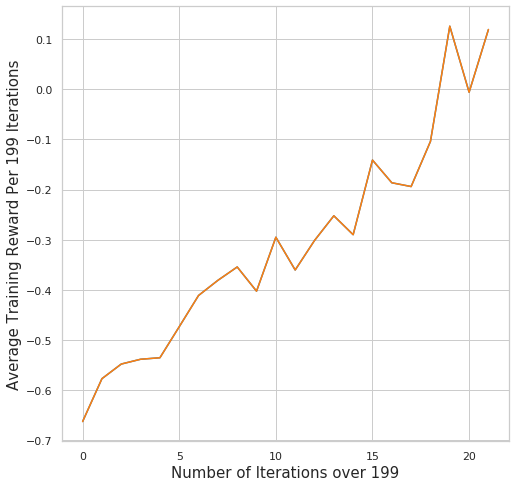}
  \caption{Trend line of average training reward.}
  \label{fig:reward}
\end{figure}

We first report experiments with $P_x = 0.7$, which show that the DQN algorithm is effective at optimizing the expected results of system decisions during training.  Figure~\ref{fig:trainLoss} shows the trend of the training loss and Figure~\ref{fig:reward} presents the reward progression during training.  Both show the system consistently improving over time.  Table \ref{tab:successes} shows the success rate over test data after each epoch. Average loss over testing data after training is $0.038$.  Both measures indicate that the model is  generalizing well to unseen examples.  Ultimately, the simulated success rate is 92\% for unseen color patches at test time.

\begin{table}[htb!]
    \centering
    \begin{tabular}{|c|c|c|}
        \hline
        & Successes & Failures\\
        \hline
       1st Epoch & 378  &  22\\
       \hline
       2nd Epoch & 369 & 31\\
       \hline
    \end{tabular}
    \caption{Success vs Failures over Test Set when following the learned policy $\pi$. The minor difference in the number of successes is because of imprecision in simulation.}
    \label{tab:successes}
\end{table}

\begin{figure}[ht!]
    \begin{tabular}{c}
        \\
        \includegraphics[width=\columnwidth]{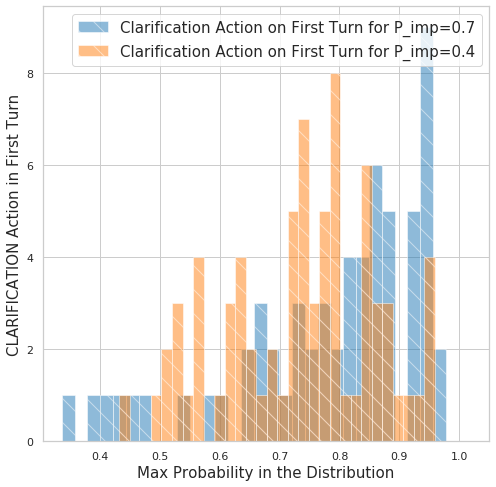}
        \\
        \includegraphics[width=\columnwidth]{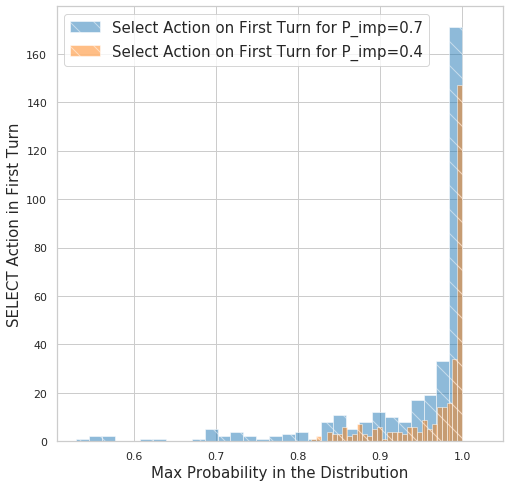}
        \\
    \end{tabular}
  \caption{A comparison of the number of \textit{Clarification} actions and \textit{Select} actions chosen by the models.}
  \label{fig:clarification}
\end{figure}

Our next experiments contrast the policies learned with $P_x=0.7$ against policies learned with $P_x=0.4$.  The differents show that the RL setup is able to learn strategies that respond appropriately to differences in simulated user behavior.  Figure \ref{fig:clarification} shows histograms of model choices at the first turn, as a function of the probability value associated with the most likely referent for the target.  Recall that the options are \textit{Selection}, which commits the system to a specific referent, and \textit{Clarification}, which aims to get more information.  We can see that the model trained with $P_{x}=0.7$---with users that are less successful in identifying the target in the first turn---chooses to ask clarification questions more often than the model trained with $P_{x}=0.4$. Thus as it was expected, the model learns to select the color patch only when it is almost certain that it can match the director's description with the target.


We observe that the model trained with $P_x=0.7$ learned to ask $66.7\%$ more clarification questions at probability values $\ge 0.8$. Similarly, it tends to select a patch $5.5\%$ less at the probability values $\ge 0.8$. This also showcases how $P_x$ controls the trade-off between clarification and selection.

\section{Related Work}

The role of the dialogue manager in dialogue systems is to interact in a natural way to help the users complete the task that the system is designed to support \cite{walker2000application}. Reinforcement learning has been traditionally successful for learning optimized yet flexible strategies in dialogue \cite{frampton2006learning,manuvinakurike2017using,walker2000application,zhang2018multimodal}. To automatically learn dialogue management strategies from dialogue data, these models rely on simulated users, as we do; see \citet{schatzmann2006survey} for a review. To learn interaction policies, we follow the approach of \citet{fang2013towards,she2017interactive}, who use a simulated human model built from the dataset (they learn grounded verb semantics interactively in situated human--robot dialogue).  



Several projects have studied deep RL for continuous state (and action) spaces, generally in robotics applications  \cite[among others]{DBLP:journals/corr/LillicrapHPHETS15}.  Nevertheless, to the best of our knowledge, our work is the first that studies the effectiveness of applying deep RL models for learning clarification strategies in dialogue.  In fact, however, because these methods make decisions based on the values of continuous parameters, they offer similar flexibility to approaches to dialogue management based on partially-observable MDPs, which are widely used in dialogue management \cite{williams2007partially,6407655}.  A comparison of partially observable MDPs to deep RL in this domain is left for future work.

Our findings provide new insights into ways that interlocutors employ disambiguation strategies while suggesting the potential discourse approaches in collaborative problem solving in dialogue. The closest work to ours is \citet{schlangen2004causes}, who presents a model of possible causes for requesting clarifications in dialogue while relating models of communication \cite{clark1996using,allwood1995activity} to the discourse semantics of \citet{asher-lascarides:2003a}.  Our work goes beyond this approach by providing tools for scaling it up, including by learning semantic models and dialogue strategies.


\section{Conclusion}

In this paper, we have described the rationale and design principles for using coherence in mixed-initiative referential communication dialogues, and described a preliminary implementation.  We showed that we could engineer significant coverage of natural human--human dialogue utterances and strategies, drawing systematically on a range of machine learning methods, and how we could fine-tune the system's purposeful behavior using reinforcement learning drawing on cognitively inspired user simulations.  At the same time, we suggested that the system's representations and architecture promise better coverage of anaphorically-mediated real-world reference and dialogues that mix agreement, partial agreement and explicit disagreement.

Some gaps issues remain to offer a complete and convincing demonstration of our framework.  We need better performance evaluations, along with closer links between the reinforcement learning models and the system's interaction with people, ultimately naive users with an intrinsic interest in communicating with the system.  We need to substantiate the capacity of the architecture to support mixed initiative by learning strategies where the system acts as the director, not just as the matcher.  We need to explore more systematic processing of discourse in context, including a robust and general treatment of anaphora and attachment.  And we need to start explicitly drawing on general methods for parsing and DRS construction.   Though we would have preferred to use domain-independent tools and resources, it remains a general challenge to build high-precision linguistic resources for natural dialogue. These diverse open problems offer exciting opportunities for future research.

\section*{Acknowledgement}

The research presented here is supported by
NSF Awards IIS-1526723 and CCF-19349243. We thank the Semdial reviewers for helpful comments, and the Mechanical
Turk annotators for their contributions.

\bibliography{anthology,acl2020}
\bibliographystyle{acl_natbib}




\end{document}